\documentclass[preprint]{elsarticle}
\journal{Journal of Rail Transport Planning and Management}
\usepackage{amsfonts, amsmath}
\usepackage{a4wide}
\usepackage{tikz}
\usepackage{graphicx} % Required for inserting images
\usepackage{amsthm} % For theorem-like environments such as definition
\usepackage{multirow}

\usepackage{bbm} % For \mathbbm
\newtheorem{definition}{Definition}
\providecommand{\tightlist}{%
  \setlength{\itemsep}{0pt}\setlength{\parskip}{0pt}}
\usetikzlibrary{arrows, arrows.meta, quotes, positioning}
\usepackage[dvipsnames]{xcolor}

\providecommand{\nextop}{^+}

\colorlet{colort1}{Violet}
\colorlet{colort2}{Peach}
\colorlet{colort3}{RubineRed}
\usepackage{todonotes}\setuptodonotes{inline,color=colort3!10!white}
\IfFileExists{xurl.sty}{\usepackage{xurl}}{} % add URL line breaks if available
\IfFileExists{bookmark.sty}{\usepackage{bookmark}}{\usepackage{hyperref}}
\hypersetup{
  pdftitle={The DISPLIB 2025 train dispatching competition},
  colorlinks=true,
  urlcolor=blue!85!black,
  linkcolor=black,
  }

\tikzstyle{sig}=[thick]
\tikzstyle{rail}=[ultra thick]

\newcommand*{\railNSend}{+ (0,.15) -- + (0,-.15) + (0,0)}

\newcommand{\trainoverarrow}[4] {
    \begin{scope}[shift={(#1,#2)}]
        \fill [color=black] (0.0, 0.25) -- (#3,0.25) -- (#3-0.15,0.5) -- (0.0,0.5) -- cycle;
        \fill [color=white] (#3-0.25,0.3) -- (#3-0.1,0.3) -- (#3-0.2,0.45) -- (#3-0.25,0.45) --  cycle;
        \draw[->,thick] (#3-0.10-#4, 0.60) -- (#3-0.10,0.60);
    \end{scope}
}
\newcommand{\trainoverarrowleft}[4] {
    \begin{scope}[shift={(#1,#2)},xscale=-1]
        \fill [color=black] (0.0, 0.25) -- (#3,0.25) -- (#3-0.15,0.5) -- (0.0,0.5) -- cycle;
        \fill [color=white] (#3-0.25,0.3) -- (#3-0.1,0.3) -- (#3-0.2,0.45) -- (#3-0.25,0.45) --  cycle;
        \draw[->,thick] (#3-0.10-#4, 0.60) -- (#3-0.10,0.60);
    \end{scope}
}

\usepackage{color}
\usepackage{fancyvrb}

\DefineVerbatimEnvironment{Highlighting}{Verbatim}{commandchars=\\\{\}}
\newenvironment{Shaded}{}{}

\newcommand{\DataTypeTok}[1]{\textcolor[rgb]{0.56,0.13,0.00}{#1}}
\newcommand{\DecValTok}[1]{\textcolor[rgb]{0.25,0.63,0.44}{#1}}

\newcommand{\ErrorTok}[1]{\textcolor[rgb]{1.00,0.00,0.00}{\textbf{#1}}}

\newcommand{\FunctionTok}[1]{\textcolor[rgb]{0.02,0.16,0.49}{#1}}

\newcommand{\OtherTok}[1]{\textcolor[rgb]{0.00,0.44,0.13}{#1}}

\newcommand{\StringTok}[1]{\textcolor[rgb]{0.25,0.44,0.63}{#1}}

\title{DISPLIB: a library of train dispatching problems}
\author{Oddvar Kloster, Bjørnar Luteberget, Carlo Mannino, Giorgio Sartor}
\date{April 2025}

\begin{document}

\begin{abstract}
Optimization-based decision support systems have a significant potential 
to reduce  delays, and thus improve efficiency
on the railways, by automatically re-routing
and re-scheduling trains after delays have occurred.
The operations research community has dedicated
a lot of effort to developing optimization algorithms for this problem, but each study is typically tightly connected
with a specific industrial use case. Code and data are seldom shared publicly. This fact hinders reproducibility, and has led to a proliferation of papers describing algorithms for more or less compatible problem definitions, without any real opportunity for readers
to assess their relative performance.
Inspired by the successful communities around  MILP, SAT, TSP, VRP, etc., we introduce a common problem definition and file format, DISPLIB, which captures all the main features of train re-routing and re-scheduling.
We have gathered problem instances from multiple real-world use cases and made them openly available.
In this paper, we describe the problem definition, the industrial instances, and a reference solver implementation.
This allows any researcher or developer to work on
the train dispatching problem without an industrial
connection, and enables the research community to perform empirical comparisons between solvers.
All materials are available online at \url{https://displib.github.io}.
\end{abstract}
\maketitle

\section{Introduction}
The trains traveling on a railway network
typically follow a strict timetable.
However, when unexpected delays happen, train dispatchers need to re-route and re-schedule the trains to prevent
delays from propagating to other trains. 
The current practice in the railway industry
is that a group of dispatchers, typically located
 in a central control room, 
handle
the decision-making manually.
This work is notoriously difficult when there are
major disturbances in the traffic, and the decisions
made in stressful situations will often be sub-optimal.

In the last ten years, digitalization of railways have
progressed to the point where 
most railway operators now have sensors and computer
systems to keep track of where trains are at any point in time, and some operators also have graphical user interfaces for managing real-time updates to timetables.
With this digital infrastructure in place, 
there is every opportunity to make use of algorithmic
decision support for dispatchers, which could alleviate some of the workload for dispatchers and 
result in more optimal re-routing and re-scheduling decisions. 
And indeed, quite a lot of academic and industrial
research efforts have gone into developing
algorithms that are able to solve train dispatching
optimization problems. 
Algorithms based on mixed-integer linear programming (MILP) are the most common, but most of the  methods from the optimization toolbox have been applied, including 
heuristics, local search, branch and bound, constraint programming, Boolean satisfiability, reinforcement learning, decompositions, and more
(see the surveys
\cite{corman2014review,fang2015survey, lamorgese2018train}).

Solving train dispatching optimization problems have proven to be hard, both  theoretically and in practice. Theoretically, the problem is NP-hard \cite{lu2004modeling}, even when limited to find a feasible solution.
Also, in terms of 
algorithm implementation, none of the scientific papers published
on this topic has been able to provide an algorithm that handles
all relevant problem instances and gives high-quality solutions in 
the short amount of time available in the real-time setting. Indeed, in industrial deployments, combinations of exact methods
and heuristics, applied on decompositions at
different time/space scales, are currently required to achieve
state-of-the-art scalability \cite{AMP25}.
Further improvements in solution methods are highly desired
by the industry, so that train dispatching
optimization can be applied more easily and in a wider range of situations.
Furthermore, the train dispatching problem also takes the role as a sub-problem problem in a host of other railway planning problems which are influenced by decisions about routing and scheduling. This includes strategic timetabling \cite{sartor2023milp}, short-term timetabling \cite{kloster2023optimization,lamorgese2017exact}, line planning \cite{zhang2021integrated}, rolling stock rotation \cite{leng2020role}, etc. All of these problems would benefit from advances in optimization techniques for the core routing and scheduling problem, i.e., the train dispatching problem.

Comparing the situation in the science of train dispatching to the communities around other optimization problems, such as the vehicle routing problem (VRP),
may provide some insight into how the academic community should organize
in order to make more progress and have more impact. VRP was
first introduced in the scientific literature in 1959 \cite{dantzig1959truck},
at which time only the tiniest problem instances could be solved with any 
expectation of producing solutions of decent quality.
Compare this with today's situation, where industrial software developers
can easily procure mature software libraries such as PyVRP and Google OR-Tools and set up solvers for highly customized problem variants, often achieving near-optimal solutions even for large-scale instances.
In the 1980s and '90s, several influential VRP researchers published
sets of hard VRP problem instances. These instances became standard benchmarks, and authors of papers presenting 
new algorithmic developments were expected to demonstrate performance
by reporting the results they had achieved on these instances.
The existence of a hard measure of algorithm performance spurred quite
a bit of healthy competition, and made it easy for researchers to demonstrate to their colleagues how good their algorithms were.

Impressive algorithmic progress has happened in several 
other communities where such  a standard for benchmarking has been central, e.g., 
mixed-integer linear programming \cite{gleixner2021miplib}, Boolean satisifability \cite{heule2024proceedings}, and traveling salesman problem \cite{reinelt1991tsplib}.
We believe that facilitating such comparisons can have a great positive impact
on scientific progress,
and that progress in train dispatching algorithms
is currently impeded by the lack of such a standard for benchmarking.
There have been some benchmarking efforts centered around competitions,
such as the Informs RAS Problem Solving Competitions \cite{ras}, the Flatland Challenge \cite{mohanty2020flatland}, and the SBB Train Schedule Optimisation Challenge \cite{sbbcompetition}.
These efforts were not aimed to create a general benchmarking framework for realistic and diverse
problem instances, but were based on specific case studies, with few problem instances,
and using more ad-hoc file formats.
And indeed, very few
papers suggesting new train dispatching methods use these 
problem instances to demonstrate algorithmic performance characteristics.

To improve on this situation, we are in this paper proposing a
problem definition called {\nobreak DISPLIB} that encompasses most important features
of train re-routing and re-scheduling problems,
while still being simple enough that minimal domain knowledge of railway operations is
required to understand it, and that algorithmic ideas can be implemented without
getting hindered by an inordinately complex problem representation.
The intention is not necessarily that industrial applications should
 use exactly this problem definition, but that the problem definition
is close enough to relevant industrial problems, so that progress in 
solving DISPLIB problems will translate into progress in train dispatching optimization in general.

We have also 
gathered a diverse set of  problem instances from  
railway infrastructure managers from multiple countries to form
the first version of the DISPLIB benchmark instance set, called DISPLIB 2025.
To create awareness and interest, we held a 
competition, running from October 2024 to April 2025, to encourage
the train dispatching research community to come up with a general algorithm
working for all the problem instances. 
Many great contributions were submitted, but only a few managed to create algorithms
that could produce feasible solutions to all the instances, and there are still many
problem instances for which no optimal solution was found.
This confirms that the DISPLIB 2025 is a challenging set of benchmark instances
that can be used by researchers in the coming years to create easily comparable and
reproducible performance results.
In the future, we intend to continue gathering problem instances
to cover a wider range of application contexts. Also, if the community
comes up with algorithms than can solve the instances
of DISPLIB 2025 to near-optimality within a short amount of time, we
also hope to gather and provide even harder problem instances to spur continued 
progress in the field in a future version of the benchmark library.
File downloads are available online at \url{https://displib.github.io},
and future updates to the set of benchmark instances and competition details will be posted there.

\section{The DISPLIB problem definition}\label{section:problem-definition}

In this section, we will define the DISPLIB routing and scheduling problem and
then show how it is suitable for modelling real-time train dispatching
problem instances.

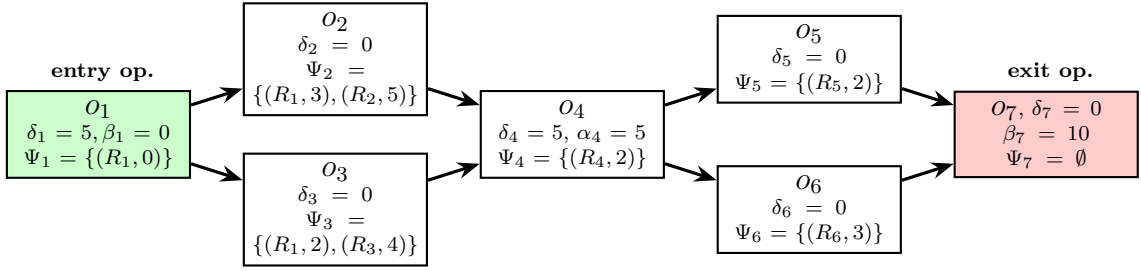
\begin{figure}
\begin{center}
\begin{tikzpicture}[auto,xscale=1.05,
                       > = Stealth, 
              box/.style = {draw=black, thick,
                            minimum height=8mm, text width=22mm, 
                            align=center,font=\footnotesize},
       every edge/.style = {draw, ->, very thick},
every edge quotes/.style = {font=\footnotesize, align=center, inner sep=1pt}
                            ]
% from bottom to top
    \node (n1) [box, fill=green!20!white] at (0, 0){{\large $o_1$} \\$\delta_1=5, \beta_1=0$\\ $\Psi_1=\left\{ (R_1, 0) \right\}$ };
    \node (n2) [box] at (3,1) {{\large $o_2$}\\ $\delta_2=0$\\ 
    $\Psi_2=\left\{ (R_1, 3), (R_2, 5) \right\}$
    };
    \node (n3) [box] at (3,-1){{\large $o_3$}\\$\delta_3=0$\\ 
    $\Psi_3=\left\{ (R_1, 2), (R_3, 4) \right\}$};
    \node (n4) [box] at (6,0) {{\large $o_4$} \\ $\delta_4=5$, $\alpha_4=5$ \\ $\Psi_4=\left\{ (R_4, 2) \right\}$};
    \node (n5) [box] at (9,1){{\large $o_5$}\\ $\delta_5=0$\\ $\Psi_5=\left\{ (R_5, 2) \right\}$};
    \node (n6) [box] at (9,-1) {{\large $o_6$}\\ $\delta_6=0$\\ $\Psi_6=\left\{ (R_6, 3)\right\}$};
    \node (n7) [box, fill=red!20!white] at (12,0){{\large $o_7$}, $\delta_7=0$ \\ $\beta_7=10$\\ $\Psi_7=\emptyset$};
\draw (n1) edge (n2)
      (n1) edge (n3) 
      (n2) edge (n4) 
      (n3) edge (n4) 
      (n4) edge (n5) 
      (n4) edge (n6) 
      (n5) edge (n7) 
      (n6) edge (n7) 
;
\node[anchor=south, font=\footnotesize] at (n1.north) {\textbf{ entry op.}};
\node[anchor=south, font=\footnotesize] at (n7.north) {\textbf{ exit op. }};
        \end{tikzpicture}
\end{center}
\caption{A train operation graph. 
The boxes represent operation $o_i$, for $i=1,\dots, 7$, labeled with a minimum duration ($\delta_i$), upper and lower bounds ($\alpha_i$, $\beta_i$) and
a set $\Psi_i$ of resources
from the set $\mathcal R = \left\{ R_1, R_2,\ldots \right\}$ to which it needs exclusive access with corresponding release times.
Some parameters are left out to simplify the presentation 
(assume $\alpha=0$, $\beta=\infty$, $\lambda=0$ where omitted).
The green node $o_1$ is the entry operation
(it has no incoming edges), and the red node labeled $o_7$ is the exit operation (it has no outgoing edges).
An example of a route for this train would be the
sequence $\pi=\langle o_1, o_2, o_4, o_6, o_7 \rangle$.
}
	\label{fig:dag}
\end{figure}

Let $I$ be a set of trains. Each train $i\in I$ performs a sequence of operations $\pi^i$ selected from a set  $O^i = \left\{ o_1^i, o_2^i, \ldots, o_{n_i}^i  \right\}$ of
$n_i$ potential operations. This sequence must start with
the \emph{entry operation}, $o_1^i\in O^i$, and end
with the \emph{exit operation}, $o_{n_i}^i$.
A successor relation $S^i \subset O^i \times O^i$ restricts
which pairs $(o^i_a,o^i_b) \in S^i$ are allowed to be immediate successors
in $\pi^i$.
$S^i$ defines a directed acyclic graph, and 
the sequence $\pi^i$ must thus be a path in this graph
starting from $o_1^i$ and ending in $o_{n_i}^i$
(see Figure~\ref{fig:dag}). Such a sequence $\pi^i$ is called a \emph{route} for the train $i$.

For each operation $o_a^i \in O^i$, 
there is a specified minimum duration $\delta_a^i \geq 0$,
a lower bound $\alpha_a^i \geq 0$, and an upper bound $\beta_a^i \geq 0$.
Given a specific route $\pi^i$,
we associate
with each 
operation $o_a^i \in \pi^i$
a \emph{start time} $t_a^i$. 
An operation ends when the next operation in the train's path starts,
i.e.,
for any pair of consecutive 
operations $o_a^i$ and $o_b^i$ in  $\pi^i$, we 
denote the successor operation $o_{a\nextop}^i=o_b^i$ and
define the 
\emph{end time} of $o_a^i$ as $t_{a\nextop}^i = t_b^i$ (for the last operation in the sequence, we define $t_{a\nextop}^i=\infty$). Then, for each $o_a^i \in \pi^i$,
we require:\[ 
\alpha_a^i \leq t_a^i \leq \beta_a^i  \quad \textrm{ and } \quad
t_a^i + \delta_a^i \leq t_{a\nextop}^i.\]
Such values of $t_a^i$ associated with each $o_a^i \in \pi^i$ define a  \emph{schedule} for the train $i$ with route $\pi^i$.

Additionally, each operation may require exclusive access to one or more resources.
%we require that a feasible solution 
%Then, a feasible solution is one that provides a feasible route and schedule for each train while %satisfying the exclusive usage of resources. 
%More formally, 
%we denote by $\vartheta(o)$ the train $i\in I$ associated with the operation $o \in O^i$, and  
We denote with $\mathcal R$ the set of all resources. Then, with each each operation $o_a^i$, we associate a set of resource requirements $\Psi_a^i = \left\{\dots,(r,\lambda),\dots \right\}$, where $r\in \mathcal R$ is a resource required by the operation and $\lambda \in \mathbbm{R}$ is a release time. The resource is occupied by the operation from the start of the operation and 
until  $\lambda$ time units after the  end of the operation.
% occupations, and with each resource occupation
% $\psi_j \in \Psi_a^i$, we associate a resource $r_j \in \mathcal R$  and a 
% release time $\lambda _j \geq 0$.

Combining the routes and schedules of all trains, we define a solution as a sequence of operations and their corresponding start times, ordered chronologically,  
 
\[ \Pi = \langle \dots, (o,t), \dots \rangle\] 
%where $t_1 \leq t_2 \leq t_3 \leq \ldots$. 

Denote by $\eta(o_{a}^i)$ the position of $o_a^i$ in the sequence $\Pi$.
If $o_a^i$ is the last operation in $\pi^i$, 
we define $\eta(o_{a\nextop}^i)=\infty$. Then we can give the following definition:
%of the 
%immediate successor of $o_a^i$ in $\pi^i$ 
%(if $o_a^i$ is the last operation in $\pi^i$, 
%we define $\eta(o_{a+1}^i)=\infty$).

\begin{definition} \textbf{Feasible solution.} 
A solution $\Pi$ is feasible if:
\begin{enumerate}
    \item For each train $i\in I$, the subsequence of 
$\Pi$ of the operations $\langle o_a^i, o_b^i, \ldots \rangle$ from $O^i$, is a route $\pi^i$, and the corresponding times $t_a^i,t_b^i,\ldots$ define a schedule for $\pi^i$;
    \item For any pair of  
operations $o_a^i,o_b^j \in \Pi$, with 
$\eta(o_a^i) < \eta (o_b^j)$ and
$i\neq j$,
and for any pair
$(r,\lambda) \in \Psi_a^i$ and $(r',\lambda') \in \Psi_b^j$ with $r=r'$, we have:
\[ \eta(o_{a\nextop}^i) < \eta(o_b^j),\quad\textrm{ and }\quad t_{a\nextop}^i+\lambda \leq t_b^j. \]
\end{enumerate}
\end{definition}

In other words, we are asking that: (1) each train is routed and scheduled correctly, independently of the other trains; (2) for every pair of operations (of different trains) that requires the same resource, then the operation that appears first in the solution must terminate, and the release time must elapse, before the other operation starts.

Note that if all the durations $\delta$ and release times $\lambda$ were strictly positive, it would not be necessary to describe the ordering of operations $\Pi$ explicitly, since any ordering by non-decreasing start times would be correct. However, to ensure correctness even when allowing instantaneous operations ($\delta=0$) or instantaneous release ($\lambda=0$), the  solution must be specified as a globally ordered sequence of operations. This can be convenient in macroscopic railway models (see Section~\ref{section:railway_as_displib}), and allows 
the problem definition to accommodate real-world problems
from a more diverse set of sources.

The cost of a solution is defined as the sum $Z(\Pi)=\sum_{c \in \mathcal C} z(c,\Pi)$ over a set $\mathcal C$ of objective components, where
$z(\cdot)$ is a measure of the delay of a component $c_j$ with respect to the current solution $\Pi$ and each component $c = (o_a^i, \bar t, \gamma, \zeta)$ defines a linear function associated with the start time $t_a^i$ of a specific operation operation $o_a^i\in O^i$, $i\in I$.
We define $z(\cdot)=0$ if the operation $o_a^i$ is not in the route $\pi^i$.
Otherwise, the measure $z(\cdot)$ is defined by a threshold
$\bar t$, after which the train is delayed and
incurs a cost $\gamma$ proportional to the
length of the delay,
or a constant cost $\zeta$, or both:

\[z(c,\Pi) = 
\begin{cases}
\gamma \cdot \max \left\{ 0, t_a^i - \overline t \right\} + \zeta \cdot H( t_a^i - \overline t ) & \text{ if } o_a^i \in \pi^i \\
0 & \text{ otherwise. }
\end{cases}
\]
Here, $H(t)$ is the step function, i.e., $H(t)=0$ if 
$t < 0$, and $H(t)=1$ if $t \geq 0$. This allows for a combination of different piecewise linear functions (see Figure \ref{fig:piecewise-cost}).
% \[H(t)=
% \begin{cases}
%     1, & \text{ if } t \geq 0,\\
%     0, & \text{ otherwise. }  
% \end{cases}
% \]

\begin{definition} \textbf{Optimal solution.} 
A solution $\Pi$ is optimal
if it feasible and the total cost $Z(\Pi)$
is the minimum among all feasible solutions.
\end{definition}

% In the first version of the DISPLIB, we considered only objective components associated with operation delays. Future versions may consider different types of objective components.

% In particular, 

A JSON file format for the input and output data used in this problem definition is described in 
\ref{appendix:displibformat}.
Note that in the JSON format, all numerical parameters are 
restricted to have integer values, to avoid any issues related to  numerical accuracy. In most railway applications, time values and 
durations are described in seconds, so using the time unit of 
1 second will typically be sufficiently precise for describing
the problem instances encountered in practice. 

\subsection{Real-time train dispatching as a DISPLIB problem}
\label{section:railway_as_displib}
The definitions above are motivated by the practical
needs to model 
real-time train dispatching problems.
We will now show an example of how to define the set of resources $\mathcal R$, and then the set of operations $O^i$ for a train $i$. 

\begin{figure}
    \centering
     \begin{tikzpicture}[xscale=0.9]
        \begin{scope}[shift={(0,0)}]
        % \node [anchor=west]at (-2,1.3) {$\phi_0$};

\draw[rail]  (1,1) \railNSend -- (3,1) -- (4,0)\railNSend -- (6,0)\railNSend -- (7,1) -- (9,1) -- (10,0)\railNSend -- (12,0)\railNSend -- 
(13,1) -- (15,1)\railNSend;

\draw[rail] (1,0)\railNSend -- (4,0);
\draw[rail] (6,0) -- (10,0);
\draw[rail] (12,0) -- (15,0)\railNSend;

    \node [anchor=south] at (2, 1)  {$R_1$};
    \node [anchor=north] at (2, 0)  {$R_2$};
    \node [anchor=north] at (5, 0)  {$R_3$};
    \node [anchor=north] at (8, 1)  {$R_4$};
    \node [anchor=north] at (8, 0)  {$R_5$};
    \node [anchor=north] at (11, 0) {$R_6$};
    \node [anchor=north] at (14, 1) {$R_7$};
    \node [anchor=north] at (14, 0) {$R_8$};
        
        %     \draw[rail] (-1,0)\railNSend -- (1.25,0) \railNSend -- (5,0) \railNSend ;
        %     \draw[rail] (1.25,0) -- (2.25,-1) -- (3,-1) -- (5,-1) \railNSend;

        % \draw[rail, ->, >=latex, dashed] (-1,0) -- (-2,0); 
        % \draw[rail, ->, >=latex, dashed] (5,0) -- (6,0);
        % \draw[rail, ->, >=latex, dashed] (5,-1) -- (6,-1);
        \trainoverarrow{1.1+0.1}{0}{0.8}{0.7} 

 \begin{scope}[opacity=0.5, transparency group]
\draw[color=colort1,line width=5pt,-{Latex[length=15pt]}] (2.2,0.375) -- (15,0.375);
\draw[color=colort1,line width=5pt] (6-0.375,0.375) -- (7-0.375,1+0.375) -- (9+0.375,1+0.375)   -- (10+0.375,0+0.375) ;
\end{scope}
        
        % \trainoverarrowleft{4.9}{0}{0.8}{0.7}
        % \draw[shorten >=0.3em, shorten <=0.3em,line width=0.5em,color=colort2] (1.25,0) -- (5,0);
        % \draw[shorten >=0.3em, shorten <=0.3em,line width=0.5em,color=colort1] (0-1,0) -- (1.25,0);
        %     \node [anchor=south] at (-0.5,0.6) {A};
        %     \node [anchor=south] at (4.5,0+0.6) {B};
        %     \node [anchor=north] at (0.25,0-0.1) {\texttt{l}};
        %     \node [anchor=north] at (3.5, 0-0.1) {\texttt{r1}};
        %     \node [anchor=north] at (3.5, -1-0.1) {\texttt{r2}};
        \end{scope}
\end{tikzpicture} 

    \caption{Railway infrastructure divided into block sections ($R_1,R_2,\ldots$) and
    a train, starting from $R_2$, traveling through the infrastructure to end up in
    $R_8$.}
    \label{fig:train}
\end{figure}
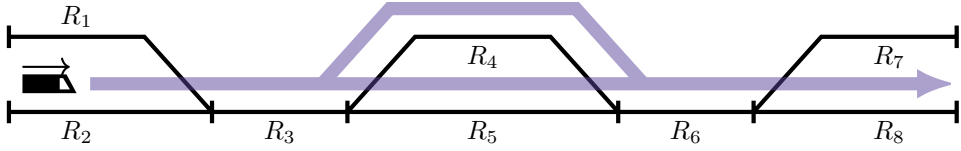

The \emph{signaling block system} (or \emph{fixed-block signaling}) is a safety system used in most railway operations around the world today. The basic principle is that the railway tracks are divided into a set of \emph{block sections}, which are sections of the infrastructure that are only allowed to be occupied
by one train  at a time. We will use resources
to represent such block sections.
% (though resources may also be used
% to model other for which trains require mutually exclusive access).
Depending on the application context, the block sections may correspond directly to the
track circuits used by the control system to
detect trains, or they may represent more approximate regions, such as 
the main and siding tracks used for meeting and overtaking, as illustrated in Figure~\ref{fig:train}.

A dispatching problem instance is limited in scope
to a dispatching region, i.e., some connected subset of a railway network.
We assume that we can disregard the part of the railway network that lies outside the dispatching region.
Now, we imagine that the train in Figure~\ref{fig:train} 
has not started traveling yet.
This means that it is currently either parked at a depot, or is traveling somewhere 
outside the dispatching region.
Assume also that the train is planned to start (e.g. from the official timetable) at some time
$\bar t_0$ in the future from the platform represented by the resource $R_2$, and
we allow the start time to be postponed.
Then, the entry operation $o_1$ has $\alpha_1=\bar t_0$ and $\beta=\infty$, and 
$\Psi_1=\left\{ (r_1, \lambda_1) \right\}$ using resource
$r_1=R_2$. The release time $\lambda_1$ represents the minimum lap between when the train starts
its next operation (which will be traveling on $R_3$), and when another train may start 
an operation occupying $R_2$, and it is often computed as the length of the train 
divided by its nominal speed. The minimum duration $\delta_1$ represents
the time that the train would use to travel through the section $R_2$.

Note that in some application contexts where a macroscopic view of
the schedule is desired, it may be reasonable to
use $\lambda_1=0$, and even $\delta_1=0$. As an example, a northbound train might arrive from a single-track section into a station at 12:00 while another southbound train is scheduled to depart from the same station at 12:00 on the same single-track section. Taken literally, these trains would need to be infinitesimally short for these events to both happen at 12:00. However, this type of model appears in real-world applications, and it would cause inconvenient cascading changes to the timetable if one would add physically accurate release times.

Continuing with the other sections, we construct operations $o_2,o_3,o_4,o_5,o_6$ for traversing 
$R_3,R_4,R_5,R_6,R_8$, respectively. Typically, $\alpha=0$ for most of the operations,
but for passenger trains there might be earliest departure times from the stations
(since it would not be allowed to leave earlier than the official timetable time), which would be applied
by setting $\alpha$ values for the operations using $R_3$ and $R_6$.

To allow the train to leave the infrastructure after $R_8$, i.e., if the train
goes to a depot or leaves the dispatching region,
 we add a dummy exit operation $o_7$ with $\Psi_7=\emptyset$.
Finally, we define the successor relation, \[S^i=\left\{
(o_1,o_2),
(o_2,o_3),
(o_2,o_4),
(o_3,o_5),
(o_4,o_5),
(o_5,o_6),
(o_6,o_7)
\right\},\]
describing the routes that the train can take. Observe that $S^i$ defines the arcs of a directed acyclic graph with vertices $O^i$, and that there is a unique node (operation) $o_1$ with no predecessors and a unique node (operation) $o_7$ with no successors.
To define the cost associated with the train, we
add an objective component $c=(o_6, \bar t, \gamma, \zeta)$ to $\mathcal C$
associated with the operation $o_6$ when the train arrives 
to its final destination. We set $\bar t$ to the official timetable
time (when the train was planned to arrive in $R_8$) and let $\gamma=1$ and $\zeta=0$.

Following this procedure for all relevant trains in the dispatching region, we get a complete
DISPLIB problem instance, but there are a few more important features of dispatching that
should be taken into account:

\begin{itemize}
    \item 

Some trains  have already started traveling in the dispatching region.
For example, if the train in Figure~\ref{fig:train} has already reached $R_4$,
then we reorganize the operations so that the operation $o_5$ corresponding to $R_4$ becomes
the new entry operation. Other operations that are not reachable through $S^i$ from $o_5$
are removed from $O^i$ and $S^i$.
The new entry operation will then have $\beta=0$ (it starts immediately), and
the minimum duration $\delta$ is reduced by an amount corresponding to the 
time that has already elapsed since the train started traveling on $R_4$.

\item 
The cost function may have a more complex dependency on the arrival
time of a train at its final destination. Piecewise
linear functions have been used in many applications (see, for example, \cite{croella2024maxsat,lamorgese2015exact}).
To implement the piecewise linear function shown
on the left in Figure~\ref{fig:piecewise-cost}, we can
use three objective components $c_1,c_2,c_3$, all 
applying to the same operation, but with different 
thresholds: 
$\bar t_1=\bar t$, $\bar t_2=\bar t+180$, $\bar t_2=\bar t+360$,
and, e.g., $\gamma_1=\gamma_2=\gamma_3=1$.
Similarly, to implement the piecewise linear function 
on the right in Figure~\ref{fig:piecewise-cost},
we use the same $\bar t$ values but let the $\gamma$ values be zero and set instead 
$\zeta_1=\zeta_2=\zeta_3=1$.

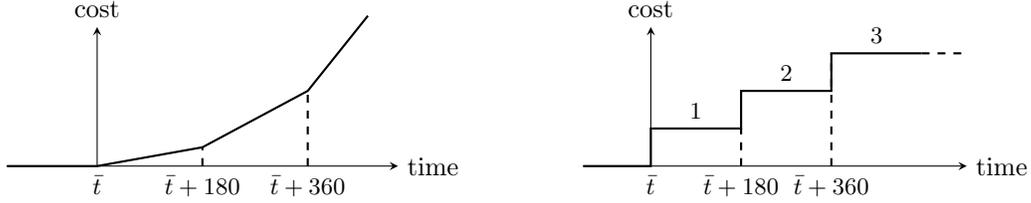
\begin{figure}
    \centering
    \begin{tikzpicture}[xscale=0.8]
    \draw[-stealth] (-1.5,0) -- (5,0) node[right]{time};
    \draw[-stealth] (0,0) -- (0,1.85) node[above]{cost};

    \draw[thick] (-1.5,0) -- (0,0) -- (1.75,0.25) -- (3.5,1) -- (4.5,2);

    \draw[thick, dashed] (0,0) node[below]{\small $\bar t $};
    \draw[thick, dashed] (1.75,0) node[below]{\small $\bar t + 180$} -- (1.75,0.25);
    \draw[thick, dashed] (3.5,0) node[below]{\small $\bar t + 360$} -- (3.5,1);
    \end{tikzpicture}\hspace{4em}
    \begin{tikzpicture}[xscale=0.6]
    \draw[-stealth] (-1.5,0) -- (7,0) node[right]{time};
    \draw[-stealth] (0,0) -- (0,1.85) node[above]{cost};
    \draw[thick] (-1.5,0) -- (0,0) node[below]{\small $\bar t $} -- (0,0.5) -- (2,0.5) -- (2,1.0) -- (4,1.0) -- (4,1.5) -- (4,1.5) -- (6,1.5);
    \draw[thick,dashed] (6,1.5) -- (7,1.5);
    \draw[thick, dashed] (1,0.5) node[above]{\small $1$};
    \draw[thick, dashed] (3,1.0) node[above]{\small $2$};
    \draw[thick, dashed] (5,1.5) node[above]{\small $3$};
    \draw[thick, dashed] (2,0) node[below]{\small $\bar t + 180$} -- (2,0.5);
    \draw[thick, dashed] (4,0) node[below]{\small $\bar t + 360$} -- (4,1.5);
\end{tikzpicture}
    \caption{Left: a convex piecewise linear function
    where cost measures delay starting ta $\bar t$ and the cost coefficient of delay increases after 180 and 360 seconds. Right: a step function where
    the delay increases when the delay surpasses 0, 180, and 360 seconds.  }
    \label{fig:piecewise-cost}
\end{figure}

\item 
Trains may be longer than a block section. 
For example, if the train in Figure~\ref{fig:train} is longer than the block section
$R_4$, then the operation that represents traveling through $R_4$ will also have a resource
occupation for $R_3$. Even worse, if the train is longer than $R_8$ and $R_6$ combined,
then traveling through $R_8$ will also need to occupy either 
$R_4$ or $R_5$, depending on the path that the train has taken.
Instead of 
creating one operation per block section, we can deal with this path dependency by creating one operation per
possible subset of resources that could be occupied when a train's head
is located at the end of a block section.
This transformation is described in more detail in 
\cite{dal20250},
and may, in the worst case, cause the set $O^i$ to become
exponentially larger. However, in practice, the relative lengths of block sections and trains
seems to result in reasonably sized sets $O^i$ (empirically, it results in at most a doubling, according to \cite{dal2022easy}).

% \end{itemize}
% There are also a few other features that are possible to model using the DISPLIB problem
% definition, but that we consider workarounds, including:
% \begin{itemize}

    \item 
    Rolling stock correspondence, i.e., the same wagons are used with a different train identity, can be implemented by joining  two trains $i,j$ into a common train with $O=O^i \cup O^j$ and $S = S^i \cup S^j \cup \left\{ (o_{n_i}^i,o_1^j) \right\}$.

\item Train correspondence constraints, i.e., a pair of trains that are forced to be in a specific station
at the same time, can be implemented by adding an additional resource that one of the trains requires
in all operations preceding the arrival to the station, and that the other train requires for leaving the station.

\item Decisions regarding cancellations (and short-turnings) can be implemented by
selecting a set of relevant operations after which the
train could cancel. Then, we add
 additional operations 
that bring the train immediately from these operations to the exit operation,
and we add  objective components for these operations with a high
cost incurred regardless of the scheduled time (the threshold is $0$).
\end{itemize}

Although the last three items of the list above could be considered less desirable workarounds,
it demonstrates the flexibility of the simple DISPLIB problem definition.
Note that, for simplicity, the infrastructure is not explicitly modelled in the DISPLIB problem definition, except through
the set of resources (for example, there is no graph structure describing the infrastructure).

\section{Problem instances}
For the DISPLIB 2025 benchmark set, 
we have gathered problem instances from four main sources.
The instances are divided into 13 categories based
on how the instances were produced from raw data. 
The categories are the following:

\begin{itemize}
    \item \texttt{smi}:
    instances received from Siemens Mobility Italy describing
    trains from a freight-dominated railway extracted from a real-time information system for train dispatching. 
    In each instance, a small set of trains were extracted
    from a larger set of trains because they represent particularly challenging interactions.
    Each train has a long sequence of operations and typically also a fairly wide range of
    routing alternatives (i.e., the graph for each train is quite complex). The
    number of trains per instances is between 5 and 13.
    Two subcategories contain different versions of these instance:
    \begin{itemize} \tightlist
  \item
    \texttt{smi\_headway}: includes non-zero release times, representing headway times
     computed from the train length and average speed.
        \item     \texttt{smi\_close}: no such headway requirement between trains.

    \end{itemize}
    
    \item \texttt{nor}: Instances received from SINTEF describing parts of the Norwegian national railway network.
    Instances were generated from a daily timetable extracted from a real-time information system for train dispatching. The timetable was modified by picking 
    a random time during the day and adding delays to 
    a random set of trains at the selected time.
    Instances were filtered
    so that only the more difficult instances are included.
    All the railway lines consists of a combination of double-track and single-track sections.
    Eight subcategories were made to distinguish between different lines and difficulty variations:
    \begin{itemize} \tightlist
        \item \texttt{nor1\_full}: describes a railway line in Norway called Jærbanen.
    The line is 120 km long with 28 stations and contains all trains for a full day.
    \item \texttt{nor1\_critical}: 
    instances generated in the same way as the previous subcategory, but where
    trains that are unlikely to be impacted by the delays are removed.
    %the trains that can be removed without changing the optimal objective value have been removed.
    \item \texttt{nor2}: describes a railway line called Gjøvikbanen.
    The line is 123 km long with 35 stations.
    The instances are limited to a 6 hour time window.
    \item \texttt{nor3}: describes a railway line called Kongsvingerbanen.
    The line is 114 km long with 26 stations.
    The  instances are limited to a 6 hour time window.
    \item
    \texttt{nor4\_small}: describes a railway line called Dovrebanen.
    The line is 492 km long with 55 stations, limited 
    to a 6 hour time window.
    \item
    \texttt{nor4\_large}: same as the previous subcategory, but
    with the full day time window.
  \item
    \texttt{nor5\_small}: 
    contains a combination of multiple lines in the Oslo region of Norway, connected
    in the Oslo Central Station, limited to a 6 hour time window.
    \item
    \texttt{nor5\_large}: same as the previous subcategory, but
    with the full day time window.
    \end{itemize}

    \item \texttt{swi}: instances based on openly  available data  from Swiss Federal Railways (SBB). 
    These instances were prepared for a computational competition \cite{sbbcompetition}.
    Instances are of increasing difficulty, ranging from a few trains to up to 467 trains per instance.

\item \texttt{wab}:
instances received from 
    Wabtec describing a set of trains in a dispatching region, 
    extracted from a real-time information system for train dispatching.
    Each of the 16 instances is made from consecutive snapshots of 
    actual train positions during a period of a few hours.
    The routing graph for each train is fairly complex, describing 
    station routing and also minimum durations in consecutive stations being 
    dependent on optional stops (i.e., a choice between stopping in one station or the next one).
    Two subcategories were made:
    \begin{itemize} \tightlist
  \item
    \texttt{wab\_large}:  all trains in the dispatching region in the next 24 hour period.
  \item \texttt{wab\_small}: same as the previous subcategory, but using a 16 hour period.
    % An additional set of 16 instances, derived from \texttt{line4\_large},
    % by keeping only the first 30 trains
    % (sorted chronologically), re-ordering (re-indexing) the trains and the
    % names of the resources (to avoid easy transferring of solutions
    % between these instances).
    \end{itemize}

%Similar to the \texttt{smi} instances, the delay thresholds of these trains are defined at time zero.

\end{itemize}

\begin{table}[]
    \centering
    \setlength{\tabcolsep}{0.2em} % for the horizontal padding

    \begin{tabular}{l c l | r c r r c r r c r}

Data &
\rule{0.7em}{0pt}&
Category & 
\rule{0.7em}{0pt}Count  &
\rule{2em}{0pt}&
\multicolumn{2}{c}{Trains}&
\rule{2em}{0pt}&
\multicolumn{2}{c}{Operations}&
\rule{1em}{0pt}&
\multicolumn{1}{c}{Resources} \\

source &
\rule{0.7em}{0pt}&
 & 
  &
&
\multicolumn{2}{c}{\small(min.-max.)}&
&
\multicolumn{2}{c}{\small(min.-max.)}&
&
\\ \hline

\multirow{2}{2em}{\texttt{smi}} && \texttt{smi\_close} & 9 && \rule{1.5em}{0pt} 5  -- & 13 && 113 -- & 1975 && 787 \\
                                && \texttt{smi\_headway} & 12 && 5  -- & 13 && 113 --  & 2245 && 790 \\
\multirow{8}{2em}{\texttt{nor}} && \texttt{nor1\_critical} & 10 && 4  -- & 16 && 148 -- & 796 && 95 \\
                                && \texttt{nor1\_full} & 10 && 40 -- & 155 && 2194 -- & 8217 && 95 \\
                                && \texttt{nor2} & 5 && 23  -- & 23 && 1448 --  & 1750 && 137 \\
                                && \texttt{nor3} & 5 && 21  -- & 22 && 1237 --  & 1325 && 79 \\
                                && \texttt{nor4\_large} & 5 && 171 -- & 365 && 17221 -- & 38125 && 408 \\                                                            && \texttt{nor4\_small} & 5 && 157 -- & 168 && 16034 -- & 17907 && 408 \\
                                && \texttt{nor5\_large} & 5 && 346 -- & 505 && 34064 -- & 50934 && 512 \\
                                && \texttt{nor5\_small} & 5 && 261 -- & 286 && 25594 -- & 28248 && 512 \\
\multirow{1}{2em}{\texttt{swi}} && \texttt{swi} & 9 && 4 -- & 467 && 326 -- & 52741 && 1301 \\
\multirow{2}{2em}{\texttt{wab}} && \texttt{wab\_large} & 16 && 44  -- & 48 && 5588 -- & 6054 && 136 \\                                                               && \texttt{wab\_small} & 16 && 30  -- & 30 && 3285 -- & 3489 && 136 \\ 
                                \end{tabular}
    \caption{Problem characteristics of the instances in the
    DISPLIB 2025, grouped by category. 
    The count column shows the number of problem instances in the category.
    The trains column shows the range of the number of trains in each
    instance (min.-max.). The operations column lists the number of operations, summed of the trains (also min.-max. over the category). The resources column shows the total number of resources
    used in the category.}
    \label{tab:instances}
\end{table}

A summary of instance characteristics is shown in Table~\ref{tab:instances}.

Together, these categories represent a wide range of situations
and represent optimization problems that are difficult enough to test the limits
of the state-of-the-art algorithm for train scheduling.
An algorithm that would produce near-optimal solutions to all of these instances in
less than 1 minute per instance would represent a significant milestone in this scientific field. 
We hope that making this collection available to the community
will spur progress toward this goal.

\section{Computational competition}

A computational competition was held
in connection with the development of the DISPLIB problem definition and file formats, and gathering of problem instances for DISPLIB 2025.
The goal of the competition was to encourage development of new train dispatching algorithms,
but also to 
(1) validate that the problem definition and file format were suitable 
for the purpose and easily understandable by the community, and
(2) make sure that the problem instances were actually hard enough to 
pose a significant challenge to the research community.

The competition was divided into two phases. Phase 1 was launched in October 2024, when 
the file format and competition rules were posted to the competition web page,
together with a 
first set of problem instances and
a solution verification program.
Participants were encouraged to submit solutions to the phase 1 instances as early as possible, to allow
the organizers to fix errors or clarify instructions. 
Then, in February 2025, phase 2 was initiated by publishing another set of
harder problem instances on the web page.

The use of existing general-purpose optimization software was allowed (including commercial solvers such as CPLEX, Gurobi, Xpress, etc.), but the use of multiple unrelated
algorithms (such as portfolio solvers) was discouraged. 
% • Offline tuning or learning phases of the algorithm are allowed, and do not count against
% the time limits or hardware limits for computing the solution. However, adaptation to
% specific problem instances are not allowed. The aim is to develop an algorithm that
% can solve arbitrary instances similar to the competition instances, and, ideally, any train
% dispatching problem instance.
Participants were required to submit a set of solutions to the problem instances
that they computed using a maximum of 10 minutes of computation time per instance.
This limit is somewhat higher than the typical real-time requirement,
which is often in the range of 1-2 minutes, in order to encourage submission of
algorithmic ideas that were
not necessarily carefully implemented or fine-tuned for real-time usage.
A rank-based scoring system was used to give points to the participants according
to their solution quality and coverage, similar to the ITC2021 competition \cite{van2023international}). 
Points are awarded to each solution based on the position among its competitors, and phase 2 instances were given more points.

In addition to the solutions to the problem instances, the participants were
required to submit a short report describing their algorithm.
An international panel of experts of railway
optimization  evaluated the submissions based on the scoring and the quality and novelty of the report. The panel selected a group of finalists
who were invited to present their algorithms at the 
 International Conference on Optimization and Decision Science (ODS) 2025.
% Lessons learned? (weakness of formula 1 ranking, surprised by the competitiveness of MILP)
More details about the competition rules and detailed results are available from the web page
\url{https://displib.github.io/}.

The competition and the gathering of problem instances validated that 
the problem definition was general enough to represent a wide range of data sources,
and the file format was understandable by the participants. Indeed, the top performing
teams had very few clarification questions to the competition organizers.
Furthermore, none of the teams managed to solve all instances to optimality, and no team outperformed all other teams on all instance categories, which confirms that the instances are challenging
and suggests that the state of the art can be improved by developing better algorithms.

\section{Reference MILP model}

% As part of the DISPLIB library, we also make available a reference MIP model. This is not meant to perform well on the DISPLIB instances, as it simply  attacks them in one of the most natural ways possible. However, it 
To provide a reference model for solving the DISPLIB problem instances, 
we present here a translation into mixed-integer linear programming (MILP).
This model is presented only to 
provide  a correct MILP formulation, which can be used for comparison, testing and inspiration.

For modeling train scheduling problems using MILP, the literature is mostly divided between big-$M$ and time-indexed formulations \cite{duarte202550}. Here we present a big-$M$ formulation, which is arguably the most direct way of modeling, and appears to be the most commonly used method in recent state-of-the-art implementations (see, for example, \cite{kloster2023optimization, leutwiler2022logic, pellegrini2019efficient}).

Reusing the notation of Section \ref{section:problem-definition}, we start by defining the continuous variables $t_a^i \in \mathbbm{R}$, $\alpha_a^i \leq t_a^i \leq \beta_a^i$, to represent the start time of operation $o_a^i\in O^i$ for train $i$, where $\alpha_a^i$ and $\beta_a^i$ are the lower bound and upper bound respectively.

First, we will focus on modeling the routing and minimum running times of the trains. 
% We assign a binary variable to each operation and to each arc of the successor graph $S^i$, representing whether they are selected in the solution. Then, we make sure the selected operations and arcs of a train correspond to a path in the train operation graph (see Figure \ref{fig:dag}). 
%
For each operation $o_a^i\in O^i$ of each train $i\in I$, we introduce a variable $x_a^i\in \{0,1\}$ that is equal to 1 if and only if the operation $o_a^i$ is part of the route $\pi^i$. Similarly, for each pair of operations $(o_a^i,o_b^i)\in S^i$, we introduce a variable $y_{a,b}^i\in \{0,1\}$ that is equal to 1 if and only if we choose $o_a^i$ and $o_b^i$ to be immediate successors in the route $\pi^i$. Then, we define $S_a^i$ as the set of successor operations of $o_a^i$, i.e., $S_a^i = \{o_b^i\in O^i : (o_a^i,o_b^i)\in S^i\}$, and with $P_a^i$ the set of predecessor operations of $o_a^i$, i.e., $P_a^i = \{o_b^i\in O^i :  (o_b^i,o_a^i)\in S^i\}$. We can now introduce the corresponding \emph{running times and routing constraints}:
\begin{equation}
    \label{eq:routing_constraints}
    \begin{array}{l}
    \begin{array}{rl}
        \sum\limits_{o_b^i\in S_1^i} y_{1,b}^i = 1&  i\in I, \\[6pt]
        \sum\limits_{o_a^i\in P_{n_i}^i} y_{a,n_i}^i = 1&  i\in I, \\[6pt]
        \sum\limits_{o_a^i\in P_c^i} y_{a,c}^i = \sum\limits_{o_b^i\in S_c^i} y_{c,b}^i&  o_c^i \in O^i\setminus \{o_1^i, o_{n_i}^i\}, i\in I, \\[6pt]
    \end{array}\\
    \left .\begin{array}{rl}
        y_{a,b}^i \leq x_a^i   \\[6pt]
        y_{a,b}^i \leq x_b^i  \\[6pt]
        x_a^i + x_b^i -1 \leq y_{a,b}^i \\[6pt]
        t_b^i - t_a^i \geq \delta_a^i \cdot y_{a,b}^i 
    \end{array}
 \right \} \quad (o_a^i,o_b^i) \in S^i, i\in I
    \end{array}
\end{equation}
The first three constraints are the classical flow conservation constraints that guarantee that we are choosing a path from entry to exit (i.e., a route), 
while the next three constraints make sure that the operations along this path are selected.
% make sure that the immediate successors 
% two successor operations are selected if and only if the 
% for each selected arc we must select the source and target operations. 
The last constraint enforces the minimum running time constraints between two subsequent operations in the chosen path\footnote{Without giving technical details which would go beyond the scope of the paper, we remark that we can avoid the use of a classical big-$M$ constraint here thanks to the the fact that the graph of operations $(O^i, S^i)$ associated with any train $i$ is acyclic.}.

We can now focus on the exclusive access to resources. We first define $O=\cup_{i\in I} O^i$ as the set of all train operations and then we introduce the set $\mathcal{A}$ of all pairs of train operations (of different trains) that require the same resource, that is:
\[\mathcal{A}=\left\{ \{o_a^i,o_b^j\}: o_a^i,a_b^j \in O, i\neq j, \exists\, (r,\lambda) \in \Psi_a^i, \exists\, (r',\lambda') \in \Psi_b^j, r=r' \right\}.\]

This is the set of all pairs of train operations that can potentially be in conflict.
Next, for every pair  $\{o_a^i,o_b^j\}\in \mathcal{A}$, we introduce the binary variable $z_{a,b}^{i,j}\in\{0,1\}$ that is equal to 1 if and only if operation $o_a^i$ precedes operation $o_b^j$. This means not only that $o_a^i$ starts before $o_b^j$, but also that $o_a^i$ finishes before $o_b^j$ starts. For example, if $z_{a,b}^{i,j} = 1$, then all the successor operations of $o_a^i$ need to start (if selected) before $o_b^j$ (plus the relevant release times).
%Note that this is not enough to guarantee a feasible DISPLIB solution $\Pi$, as it is not clear how to break ties whenever $t_{a+1}^i = t_b^j$ and the order in which the operations are sequenced in $\Pi$ is important. There are several ways to deal with events that happen at the same time. Here we present one of them, which is to slightly redefine the precedence variable $z_{a,b}^{i,j}$ to be equal to 1 if and only if operation $o_a^i$ ends strictly before operation $o_b^j$ starts, i.e., $t_{a+1}^i < t_b^j$. 
Then, 
%for each pair of operations $(o_a^i,o_b^j) \in \mathcal{A}$ and for each pair of resource occupations  $((r,\lambda),(p,\vartheta))\in \Psi_a^i\times \Psi_b^j$ such that $r=p$, 
we can write the following \emph{pairwise resource constraints}:
\begin{equation}
    \label{eq:resource_constraints}
    \begin{array}{l}
\left .\begin{array}{rl}
        t_{\bar a}^i - t_b^j \leq - \lambda + M(1 - z_{a,b}^{i,j})& o_{\bar a}^i \in S_a^i,  (r,\lambda) \in \Psi_a^i,  \exists\, (r,\cdot)\in \Psi_b^i \\[6pt]
        t_{\bar b}^j - t_a^i \leq - \lambda' + M(1 - z_{b,a}^{i,j})& o_{\bar b}^j \in S_b^i, (r',\lambda ') \in \Psi_b^i,  \exists\, (r',\cdot)\in \Psi_a^i  \\[6pt]
                x_a^i  \geq z_{a,b}^{i,j} + z_{b,a}^{i,j}\\[6pt]
        x_b^i \geq z_{a,b}^{i,j} + z_{b,a}^{i,j}\\[6pt]
        x_a^i + x_b^i - 1 \leq z_{a,b}^{i,j} + z_{b,a}^{i,j}
    \end{array}
        \right\} \quad \{o_a^i,o_b^j\} \in \mathcal{A}.\\

     \end{array}
\end{equation}

The first constraint says that if operation $x_a^i$ precedes operation $x_b^j$ and the release time of the common resource for operation $o_a^i$ is $\lambda$, then we have $t_c^i + \lambda \leq t_b^j$ for all successor operations $o_c^i$ of $o_a^i$. Similarly the second. 
%The third constraint makes sure that at most one operation can have precedence onto the other, and 
The third enforces the precedence decision when both operations are selected, whereas the last two ensure that no precedence is impose when at least one of the two operations is not selected. 

Note that while the previous constraints guarantee exclusive use of common resource for each pair of corresponding train operations, this is in general not enough to guarantee the feasibility of a solution. The typical example involves two trains traveling in opposite directions and two resources, where train $A$ traverses first resource $r_1$ (in 10 units of time) and then $r_2$ while train $B$ traverses first $r_2$ (in 10 units of time) and then $r_1$. Assuming that the release times are equal to 0, a schedule with $t_{r_1}^A=0,t_{r_2}^A=10$ and $t_{r_2}^B=0,t_{r_1}^B=10$ would be feasible according the constraints we defined so far (here we omitted the concept of train operations for simplicity of presentation). Sometimes, this is called \emph{swapping}, as the trains are instantaneously swapping the usage of resources at time 10. 

The swapping behavior is prevented by imposing a global ordering of operations. 
%We can apply the same idea to our model. 
For each operation $o_a^i\in O$, we introduce a non-negative integer variable $u_a^i$. Then, we simply order all operations based on all their precedence relationships by introducing the following \emph{ordering constraints}:
\begin{equation}
    \label{eq:ordering_constraints}
    \begin{array}{l}
    u_a^i - u_b^i \leq - 1 + M(1 - y_{a,b}^i) \qquad (o_a^i,o_b^i) \in S^i, i\in I,  \\[6pt]
   \left . \begin{array}{rl}
        u_{\bar a}^i - u_b^j \leq - 1 + M(1 - z_{a,b}^{i,j})& o_{\bar a}^i\in S_a^i \\
        u_{\bar b}^j - u_a^i \leq - 1 + M(1 - z_{b,a}^{i,j})& o_{\bar b}^j\in S_b^j
    \end{array}\right\} 
     \quad \{o_a^i,o_b^j\} \in \mathcal{A}.
     \end{array}
\end{equation}

Lastly, we need to model the objective function. For each objective component $c=(o_a^i,\overline{t},\gamma,\zeta)\in \mathcal{C}$, we define a binary variable $v_c$ that is equal to 1 if  $t_a^i - \overline{t} \geq 0$. We also define a nonnegative continuous variable $w_c\in\mathbbm{R}_+$ to model the cost of each component $c$. Then, we model the objective components with the following \emph{objective constraints}:
\begin{equation}
    \label{eq:objective_constraints}
    \begin{array}{rl}
        t_a^i - \overline{t} \leq M v_c& c=(o_a^i, \overline{t}, \gamma, \zeta)\in \mathcal{C}, \\ [6pt]
        %t_a^i - \overline{t} \geq -M (1 - v_c)& c=(o_a^i, \overline{t}, \gamma, \zeta)\in \mathcal{C},\\ [6pt]
        w_c \geq \gamma (t_a^i - \overline{t}) + \zeta v^c& c=(o_a^i, \overline{t}, \gamma, \zeta)\in \mathcal{C}.\\ [6pt]
    \end{array}
\end{equation}
The first constraint makes sure that $v_c = 1$ when $t_a^i \geq \overline{t}$, while the second constraint models the sum of the linear component with the step component.

The full MILP formulation then consists of minimizing $\sum_{c\in C} w_c$ subject to constraints \eqref{eq:routing_constraints} to \eqref{eq:objective_constraints}. A DISPLIB solution can then be retrieved from the solution of the MILP model by taking all operations $o_a^i\in O$ with $x_a^i=1$ and sorting them  by $(t_a^i, u_a^i)$, lexicographically. 

\section{Conclusions and future work}

This paper presents 
the DISPLIB train dispatching problem,
and a set of challenging benchmark problem instances
derived from real-world train dispatching applications.
The problem instances are freely available online, and are accompanied
by a verification program and a reference MILP solver implementation.
The problem statement and the hardness of the instances has been validated
through a computational competition held in 2024-2025.
We hope that this benchmarking framework will spur innovation and
interest in the field of train dispatching algorithms, 
and provide an infrastructure for  fruitful research advancing the state of the art.

We intend the DISPLIB benchmark set to be an ongoing effort, and
we are currently gathering additional problem instances from dispatching applications
from multiple countries. 
Any researchers or developers of train dispatching
optimization applications who have data describing relevant problem instances are welcome
to contribute them to the authors for inclusion into
a future version of the DISPLIB benchmark set.

% We are currently in the process of gathering data and adapting 
% representations for additional instances from dispatching applications from Spain, Germany, the Netherlands.
% These will be used in an upcoming next version of the DISPLIB
% benchmark set.
% Any researchers or developers of train dispatching
% optimization applications who have data describing relevant problem instances are welcome
% to contribute them to the authors for inclusion into
% a future version of the DISPLIB benchmark set.

\section*{Acknowledgements}
\noindent Thanks to SINTEF Digital, Siemens Mobility Italy, Wabtec Corporation, and Swiss Federal Railways (SBB)
%\footnote{See \href{https://data.sbb.ch/}{https://data.sbb.ch}}
for providing/publishing raw data for the
problem instances.
Thanks to the organizers of the International Conference on Optimization and Decision Science (ODS) for
supporting and publicizing the competition. 
Thanks also to the members of the scientific committee of the computational competition, 
Giorgio Sartor,
Marcella Samà,
Paolo Ventura,
Steven Harrod, and
Dennis Huisman.
This work was partially funded by the European Union through the project MOTIONAL, Grant Agreement 101101973. Views and opinions expressed are however those of
the author(s) only and do not necessarily reflect those of the European Union or the
Europe’s Rail Joint Undertaking. Neither the European Union nor the Europe’s Rail Joint Undertaking
can be held responsible for them.  
 
\bibliographystyle{plain}
\bibliography{bibs}

\appendix

\clearpage
\section{The DISPLIB file format}
\label{appendix:displibformat}

% \hypertarget{json-format}{%
% \subsection{JSON format}\label{json-format}}

The DISPLIB problem instances are stored as standard JSON files.
The JSON object for a DISPLIB problem instance contains two keys:
\texttt{trains} and \texttt{objective}, giving the following overall
structure of the JSON file, where \texttt{...} is a placeholder:

\begin{Shaded}
\begin{Highlighting}[]
\FunctionTok{\{} \DataTypeTok{"trains"}\FunctionTok{:} \OtherTok{[[}\FunctionTok{\{} \ErrorTok{...} \ErrorTok{operation} \ErrorTok{...} \FunctionTok{\}}\OtherTok{,} \ErrorTok{...} \OtherTok{],} \ErrorTok{...} \OtherTok{]}\FunctionTok{,}
  \DataTypeTok{"objective"}\FunctionTok{:} \OtherTok{[}\FunctionTok{\{} \ErrorTok{...} \ErrorTok{component} \ErrorTok{...} \FunctionTok{\}}\OtherTok{,} \ErrorTok{...} \OtherTok{]} \FunctionTok{\}}
\end{Highlighting}
\end{Shaded}

\hypertarget{trains}{%
\subsection{Trains}\label{trains}}

The top-level \texttt{trains} key contains a list of trains, where each
train is a list of operations. References to specific trains are given
as a zero-based index into the \texttt{trains} list, and references to
operations are given as a zero-based index into a specific train's list
of operations. Each operation is a JSON object with the following keys:

\begin{itemize}
\tightlist
\item
  \textbf{\texttt{start\_lb}} (optional, number): the earliest start
  time of the operation. Must be a non-negative integer. If the key is
  not present, defaults to 0.
\item
  \textbf{\texttt{start\_ub}} (optional, number): the latest start time
  of the operation. Must be a non-negative integer. If the key is not
  present, defaults to infinity (i.e., no upper bound).
\item
  \textbf{\texttt{min\_duration}} (number): the minimum duration, i.e.,
  the minimum time between the start time and the end time of the
  operation. Must be a non-negative integer.
\item
  \textbf{\texttt{resources}} (optional, list): a list of resources used
  by the train while performing the operation. If the key is not
  present, defaults to the empty list. Each resource usage is given as a
  JSON object with the following keys:

  \begin{itemize}
  \tightlist
  \item
    \textbf{\texttt{resource}} (string): the name of a resource.
  \item
    \textbf{\texttt{release\_time}} (optional, number): the release time for
		  the resource, i.e., the minimum duration between the end time
		  of this operation and the start time of any subsequent
		  operation (of a different train) using the same resource.  If
		  the key is not present, defaults to 0.
  \end{itemize}
\item
  \textbf{\texttt{successors}} (list): a list of alternative successor
  operations, given as zero-based indices into the list of this train's
  operation. The list must be non-empty unless this operation is the
  \emph{exit operation}.
\end{itemize}

For example, an operation that starts between time 7 and 8 and has a
duration of 5 time units, uses the resource \texttt{R1}, and releases it
immediately after the end event, and must be succeeded by the operation
with index 2, would be formatted as follows:

\begin{Shaded}
\begin{Highlighting}[]
\FunctionTok{\{} \DataTypeTok{"start\_lb"}\FunctionTok{:} \DecValTok{7}\FunctionTok{,} \DataTypeTok{"start\_ub"}\FunctionTok{:} \DecValTok{8}\FunctionTok{,} \DataTypeTok{"min\_duration"}\FunctionTok{:} \DecValTok{5}\FunctionTok{,}
  \DataTypeTok{"resources"}\FunctionTok{:} \OtherTok{[}\FunctionTok{\{} \DataTypeTok{"resource"}\FunctionTok{:} \StringTok{"R1"} \FunctionTok{\}}\OtherTok{]}\FunctionTok{,} \DataTypeTok{"successors"}\FunctionTok{:} \OtherTok{[}\DecValTok{2}\OtherTok{]} \FunctionTok{\}}
\end{Highlighting}
\end{Shaded}

For each train, the list of operations is ordered topologically, i.e., 
all successors for an operation appear after it in the list. 
Note that this also means that the entry operation will always be at index 0,
and the exit operation will always be the last operation in the list.

\hypertarget{objective}{%
\subsection{Objective}\label{objective}}

The top-level \texttt{objective} key contains a list of objective
components. Each objective component is a JSON object containing the key
\texttt{type} for determining an objective component type, and other
keys depending on the type. Only one type, called \textbf{operation
delay} (see Sec. 2.1), is defined in DISPLIB 2025 (the \texttt{type} key
is included for forward compatibility).

The operation delay objective component is decribed as a JSON object
with the following keys:

\begin{itemize}
\tightlist
\item
  \textbf{\texttt{type}} (string): must contain the string
  \texttt{"op\_delay"}.
\item
  \textbf{\texttt{train}} (number): reference to a train as an index
  into the top-level \texttt{trains} list.
\item
  \textbf{\texttt{operation}} (number): reference to an operation as an
  index into the list defining the train's operation graph.
\item
  \textbf{\texttt{threshold}} (number): a time after which this delay
  component is activated, as defined in the formula above. If the key is
  not present, defaults to 0.
\item
  \textbf{\texttt{increment}} (number): the constant \(d\) in the
  formula above. Must be a non-negative integer. If the key is not
  present, defaults to 0.
\item
  \textbf{\texttt{coeff}} (number): the constant \(c\) in the formula
  above. Must be a non-negative integer. If the key is not present,
  defaults to 0.
\end{itemize}

For example, an objective component that measures the time that train
0's operation 5 is delayed beyond time 10, would be formatted as
follows:

\begin{Shaded}
\begin{Highlighting}[]
\FunctionTok{\{} \DataTypeTok{"type"}\FunctionTok{:} \StringTok{"op\_delay"}\FunctionTok{,} \DataTypeTok{"train"}\FunctionTok{:} \DecValTok{0}\FunctionTok{,} \DataTypeTok{"operation"}\FunctionTok{:} \DecValTok{5}\FunctionTok{,}
  \DataTypeTok{"threshold"}\FunctionTok{:} \DecValTok{10}\FunctionTok{,} \DataTypeTok{"coeff"}\FunctionTok{:} \DecValTok{1} \FunctionTok{\}}
\end{Highlighting}
\end{Shaded}

\hypertarget{solution-format}{%
\subsection{Solution format}\label{solution-format}}

Solutions to a DISPLIB problem are given as a JSON object (typically in
a separate file from the problem instance), containing the following
keys:

\begin{itemize}
\tightlist
\item
  \textbf{\texttt{objective\_value}} (number): is the objective value of
  the solution (according to the \texttt{objective} from the problem
  instance).
\item
  \textbf{\texttt{events}} (list): is an ordered list of start events,
  i.e.~a list of references to operations to be started in the given
  order. Each start event is described by a JSON object with the
  following keys:

  \begin{itemize}
  \tightlist
  \item
    \textbf{\texttt{time}} (number): the time at which to start the
    operation. Must be a non-negative integer.
  \item
    \textbf{\texttt{train}} (number): reference to a train as an index
    into the top-level \texttt{trains} list.
  \item
    \textbf{\texttt{operation}} (number): reference to an operation as
    an index into the list defining the train's operation graph.
  \end{itemize}
\end{itemize}

\hypertarget{an-example-problem}{%
\subsection{An example problem}\label{an-example-problem}}

Let's consider two trains meeting at a junction, coming from opposite
directions. Train A is currently occupying the left part of the track
(resource \texttt{L}), and may proceed either to the upper right track
(resource \texttt{R1}), which train B is currently occupying, or to the
lower right track (resource \texttt{R2}), which is currently free.

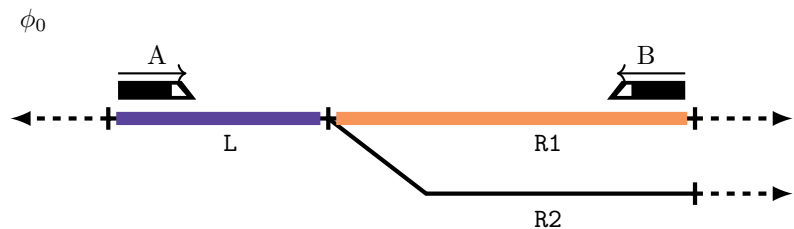
\begin{center} 
\vspace{0.5em} \begin{tikzpicture}[xscale=1.3]
        \begin{scope}[shift={(0,0)}]
        \node [anchor=west]at (-2,1.3) {$\phi_0$};
            \draw[rail] (-1,0)\railNSend -- (1.25,0) \railNSend -- (5,0) \railNSend ;
            \draw[rail] (1.25,0) -- (2.25,-1) -- (3,-1) -- (5,-1) \railNSend;

        \draw[rail, ->, >=latex, dashed] (-1,0) -- (-2,0); 
        \draw[rail, ->, >=latex, dashed] (5,0) -- (6,0);
        \draw[rail, ->, >=latex, dashed] (5,-1) -- (6,-1);
        \trainoverarrow{-1+0.1}{0}{0.8}{0.7} \trainoverarrowleft{4.9}{0}{0.8}{0.7}
        \draw[shorten >=0.3em, shorten <=0.3em,line width=0.5em,color=colort2] (1.25,0) -- (5,0);
        \draw[shorten >=0.3em, shorten <=0.3em,line width=0.5em,color=colort1] (0-1,0) -- (1.25,0);
            \node [anchor=south] at (-0.5,0.6) {A};
            \node [anchor=south] at (4.5,0+0.6) {B};
            \node [anchor=north] at (0.25,0-0.1) {\texttt{L}};
            \node [anchor=north] at (3.5, 0-0.1) {\texttt{R1}};
            \node [anchor=north] at (3.5, -1-0.1) {\texttt{R2}};
        \end{scope}
\end{tikzpicture} \vspace{0.5em} 
\end{center}

%\clearpage
The operation graphs for the two trains would look as follows:

\begin{figure}[h!]
\begin{center}
\begin{tikzpicture}[auto,xscale=1.2, yscale=0.5,
                       > = Stealth, 
              box/.style = {draw=black, thick,
                            minimum height=8mm, text width=22mm, 
                            align=center,font=\footnotesize},
       every edge/.style = {draw, ->, very thick},
every edge quotes/.style = {font=\footnotesize, align=center, inner sep=1pt}
                            ]
% from bottom to top
    \node (na0) [box, fill=colort1!20!white] at (0, 0){op 0, $\delta=5$\\$t=0$\\ res.: $[$ \texttt{L} $]$};
    \node (na1) [box] at (3,1) {op 1, $\delta=5$\\ res.: $[$ \texttt{R1} $]$};
    \node (na2) [box] at (3,-1){op 2, $\delta=5$\\ res.: $[$ \texttt{R2} $]$};
    \node (na3) [box, fill=gray!10!white] at (6,0){op 3, $\delta=0$ \\ res.: $[]$};
\draw (na0) edge (na1)
      (na0) edge (na2) 
      (na1) edge (na3)
      (na2) edge (na3)
;
\node[anchor=south west] at (na0.north west) {\textbf{Train A}};

\begin{scope}[shift={(0,-4)}]
    \node (nb0) [box, fill=colort2!20!white] at (0, 0){op 0, $\delta=5$\\$t=0$\\ res.: $[$ \texttt{R1} $]$};
    \node (nb1) [box] at (3,0) {op 1, $\delta=5$\\ res.: $[$ \texttt{L} $]$};
    \node (nb2) [box, fill=gray!10!white] at (6,0){op 2, $\delta=0$ \\ res.: $[]$};
\draw (nb0) edge (nb1)
      (nb1) edge (nb2) 
;
\node[anchor=south west] at (nb0.north west) {\textbf{Train B}};
\end{scope}
\end{tikzpicture}
\end{center}
% \caption{....}
\end{figure}

The objective is to minimize the time when train B finishes traveling
through \texttt{L}. The JSON problem instance for this problem is:

\begin{Shaded}
\begin{Highlighting}[]
\FunctionTok{\{}
  \DataTypeTok{"trains"}\FunctionTok{:} \OtherTok{[}
    \OtherTok{[}\FunctionTok{\{} \DataTypeTok{"start\_ub"}\FunctionTok{:} \DecValTok{0}\FunctionTok{,}
       \DataTypeTok{"min\_duration"}\FunctionTok{:} \DecValTok{5}\FunctionTok{,}
       \DataTypeTok{"resources"}\FunctionTok{:} \OtherTok{[}\FunctionTok{\{} \DataTypeTok{"resource"}\FunctionTok{:} \StringTok{"L"} \FunctionTok{\}}\OtherTok{]}\FunctionTok{,}
       \DataTypeTok{"successors"}\FunctionTok{:} \OtherTok{[}\DecValTok{1}\OtherTok{,} \DecValTok{2}\OtherTok{]} \FunctionTok{\}}\OtherTok{,}
     \FunctionTok{\{} \DataTypeTok{"min\_duration"}\FunctionTok{:} \DecValTok{5}\FunctionTok{,}
       \DataTypeTok{"successors"}\FunctionTok{:} \OtherTok{[}\DecValTok{3}\OtherTok{]}\FunctionTok{,}
       \DataTypeTok{"resources"}\FunctionTok{:} \OtherTok{[}\FunctionTok{\{} \DataTypeTok{"resource"}\FunctionTok{:} \StringTok{"R1"} \FunctionTok{\}}\OtherTok{]}\FunctionTok{\}}\OtherTok{,}
     \FunctionTok{\{} \DataTypeTok{"min\_duration"}\FunctionTok{:} \DecValTok{5}\FunctionTok{,}
       \DataTypeTok{"successors"}\FunctionTok{:} \OtherTok{[}\DecValTok{3}\OtherTok{]}\FunctionTok{,}
       \DataTypeTok{"resources"}\FunctionTok{:} \OtherTok{[}\FunctionTok{\{} \DataTypeTok{"resource"}\FunctionTok{:} \StringTok{"R2"} \FunctionTok{\}}\OtherTok{]}\FunctionTok{\}}\OtherTok{,}
     \FunctionTok{\{} \DataTypeTok{"min\_duration"}\FunctionTok{:} \DecValTok{0}\FunctionTok{,}
       \DataTypeTok{"successors"}\FunctionTok{:} \OtherTok{[]}\FunctionTok{\}}\OtherTok{],}
    \OtherTok{[}\FunctionTok{\{} \DataTypeTok{"start\_ub"}\FunctionTok{:} \DecValTok{0}\FunctionTok{,}
       \DataTypeTok{"min\_duration"}\FunctionTok{:} \DecValTok{5}\FunctionTok{,}
       \DataTypeTok{"resources"}\FunctionTok{:} \OtherTok{[}\FunctionTok{\{} \DataTypeTok{"resource"}\FunctionTok{:} \StringTok{"R1"} \FunctionTok{\}}\OtherTok{]}\FunctionTok{,}
       \DataTypeTok{"successors"}\FunctionTok{:} \OtherTok{[}\DecValTok{1}\OtherTok{]}\FunctionTok{\}}\OtherTok{,}
     \FunctionTok{\{} \DataTypeTok{"min\_duration"}\FunctionTok{:} \DecValTok{5}\FunctionTok{,}
       \DataTypeTok{"resources"}\FunctionTok{:} \OtherTok{[}\FunctionTok{\{} \DataTypeTok{"resource"}\FunctionTok{:} \StringTok{"L"} \FunctionTok{\}}\OtherTok{]}\FunctionTok{,}
       \DataTypeTok{"successors"}\FunctionTok{:} \OtherTok{[}\DecValTok{2}\OtherTok{]}\FunctionTok{\}}\OtherTok{,}
     \FunctionTok{\{} \DataTypeTok{"min\_duration"}\FunctionTok{:} \DecValTok{0}\FunctionTok{,}
       \DataTypeTok{"successors"}\FunctionTok{:} \OtherTok{[]}\FunctionTok{\}}\OtherTok{]]}\FunctionTok{,}
  \DataTypeTok{"objective"}\FunctionTok{:} \OtherTok{[}\FunctionTok{\{} \DataTypeTok{"type"}\FunctionTok{:} \StringTok{"op\_delay"}\FunctionTok{,}
                  \DataTypeTok{"train"}\FunctionTok{:} \DecValTok{1}\FunctionTok{,}
                  \DataTypeTok{"operation"}\FunctionTok{:} \DecValTok{2}\FunctionTok{,}
                  \DataTypeTok{"coeff"}\FunctionTok{:} \DecValTok{1}\FunctionTok{\}}\OtherTok{]}
\FunctionTok{\}}
\end{Highlighting}
\end{Shaded}

The following JSON object is a feasible (and optimal) solution to the
problem instance:

\begin{Shaded}
\begin{Highlighting}[]
\FunctionTok{\{} \DataTypeTok{"objective\_value"}\FunctionTok{:} \DecValTok{10}\FunctionTok{,} \DataTypeTok{"events"}\FunctionTok{:} \OtherTok{[}
  \FunctionTok{\{}\DataTypeTok{"time"}\FunctionTok{:} \DecValTok{0}\FunctionTok{,} \DataTypeTok{"train"}\FunctionTok{:} \DecValTok{0}\FunctionTok{,} \DataTypeTok{"operation"}\FunctionTok{:} \DecValTok{0}\FunctionTok{\}}\OtherTok{,}
  \FunctionTok{\{}\DataTypeTok{"time"}\FunctionTok{:} \DecValTok{0}\FunctionTok{,} \DataTypeTok{"train"}\FunctionTok{:} \DecValTok{1}\FunctionTok{,} \DataTypeTok{"operation"}\FunctionTok{:} \DecValTok{0}\FunctionTok{\}}\OtherTok{,}
  \FunctionTok{\{}\DataTypeTok{"time"}\FunctionTok{:} \DecValTok{5}\FunctionTok{,} \DataTypeTok{"train"}\FunctionTok{:} \DecValTok{0}\FunctionTok{,} \DataTypeTok{"operation"}\FunctionTok{:} \DecValTok{2}\FunctionTok{\}}\OtherTok{,}
  \FunctionTok{\{}\DataTypeTok{"time"}\FunctionTok{:} \DecValTok{5}\FunctionTok{,} \DataTypeTok{"train"}\FunctionTok{:} \DecValTok{1}\FunctionTok{,} \DataTypeTok{"operation"}\FunctionTok{:} \DecValTok{1}\FunctionTok{\}}\OtherTok{,}
  \FunctionTok{\{}\DataTypeTok{"time"}\FunctionTok{:} \DecValTok{10}\FunctionTok{,} \DataTypeTok{"train"}\FunctionTok{:} \DecValTok{1}\FunctionTok{,} \DataTypeTok{"operation"}\FunctionTok{:} \DecValTok{2}\FunctionTok{\}}\OtherTok{,}
  \FunctionTok{\{}\DataTypeTok{"time"}\FunctionTok{:} \DecValTok{10}\FunctionTok{,} \DataTypeTok{"train"}\FunctionTok{:} \DecValTok{0}\FunctionTok{,} \DataTypeTok{"operation"}\FunctionTok{:} \DecValTok{3}\FunctionTok{\}}\OtherTok{]} \FunctionTok{\}}
\end{Highlighting}
\end{Shaded}

Note that this is an example where the global ordering of events is
important beyond the ordering implied by the time values: if we switch
the ordering of the two events at time 5, the result is not a feasible
solution since operation 1 of train B allocates \texttt{L} while it is
still in use by train A.

\end{document}